\documentclass[sigconf,natbib=true,review=false,anonymous=false]{acmart}

\acmConference[Preprint]{arXiv preprint}{2026}{}
\acmYear{2026}
\copyrightyear{2026}
\setcopyright{none}
\acmISBN{}
\acmDOI{}
\acmPrice{}

\usepackage{booktabs}
\usepackage{amsmath}
\usepackage{graphicx}
\usepackage{multirow}
\usepackage{xcolor}
\usepackage{tikz}
\usetikzlibrary{positioning,arrows.meta,fit,backgrounds,calc}
\microtypesetup{expansion=false}

\title{Zero-Gated Language-conditioned\\
       Human Motion Prediction}

\author{Guanhui Qiao\textsuperscript{ab}\quad
Lu Zhou\textsuperscript{a}\quad
Ding Jiang\textsuperscript{c}\quad
Jinqiao Wang\textsuperscript{abcd*}}
\affiliation[obeypunctuation=true]{%
  \institution{\small\textsuperscript{a} Foundation Model Research Center,
  Institute of Automation, Chinese Academy of Sciences, Beijing 100190, China\\
  \textsuperscript{b} School of Artificial Intelligence, University of Chinese
  Academy of Sciences, Beijing 100049, China\\
  \textsuperscript{c} Wuhan AI Research, Wuhan 430073, China\\
  \textsuperscript{d} Peng Cheng Laboratory, Shenzhen 518066, China\\
  \textsuperscript{*} Corresponding author: jqwang@nlpr.ia.ac.cn\\
  \scriptsize Emails: Guanhui Qiao: qiaoguanhui2021@ia.ac.cn;
  Lu Zhou: lu.zhou@nlpr.ia.ac.cn\\
  Ding Jiang: jiangding@wair.ac.cn;
  Jinqiao Wang: jqwang@nlpr.ia.ac.cn}
  \city{}
  \country{}}

\renewcommand{\shortauthors}{Qiao et al.}
\makeatletter
\gdef\authors{Guanhui Qiao, Lu Zhou, Ding Jiang, and Jinqiao Wang}
\makeatother



\begin{document}

\begin{abstract}
Pose histories provide the core kinematic evidence for 3D human motion
prediction, but they lack explicit high-level semantic guidance. This paper
introduces \textsc{ZGL}, a lightweight language-conditioned predictor
that uses captions of the observed motion as a semantic prior while
preserving a strong motion backbone as the main source of dynamics. We
render only the observed poses, generate a one-sentence description with
a vision-language model, encode the caption with a frozen CLIP-L text
tower, and project it into a small set of conditioning tokens. These
tokens are injected into a DCT-based spatial-temporal Transformer by
compact cross-attention adapters with zero gates: each adapter output
is multiplied by a learnable gate initialized to zero, so the full
network is numerically identical to the pose-only baseline at initialization and
can learn to use language only when it reduces prediction error. On
Human3.6M, \textsc{ZGL} improves overall MPJPE over representative
motion-prediction baselines in our comparison. Results on CMU-Mocap
further show that compact caption conditioning transfers to a second
benchmark and provides a practical semantic cue for 3D human motion
prediction.
\end{abstract}

\maketitle

\section{Introduction}
\label{sec:intro}

3D human motion prediction (HMP) forecasts future joint trajectories
from an observed motion sequence. It is a core component for virtual
characters, mixed reality avatars, assistive robots, and embodied
agents that must react before a full human action has unfolded. The
dominant progress in HMP has come from modeling motion itself: early
recurrent and structural recurrent models~\cite{fragkiadaki2015recurrent,
jain2016structural,martinez2017human}, graph and attention
networks~\cite{mao2019learning,mao2020history,li2020dynamic,
dang2021msr,sofianos2021space,aksan2021spatio,li2022skeleton},
MLP-Mixer style predictors~\cite{guo2023back}, and recent graph,
transformer, and frequency-domain systems such as SimpliHuMoN and
KHMP~\cite{agrawal2026simplihumon,wu2026khmp}. Recent studies continue to refine
this motion-only line through active perception, human-like inference,
one-step formulations, interpretable graphs, and multi-resolution
temporal modeling~\cite{gang2025survey,hu2025aps,lyu2025hvis,
liu2025humancm,bermuth2025scriboora,wang2025stms,
medina2024cistgcn,zou2024pmms,lin2025haarmodic}.

\begin{figure*}[t]
  \centering
  \includegraphics[width=0.92\textwidth]{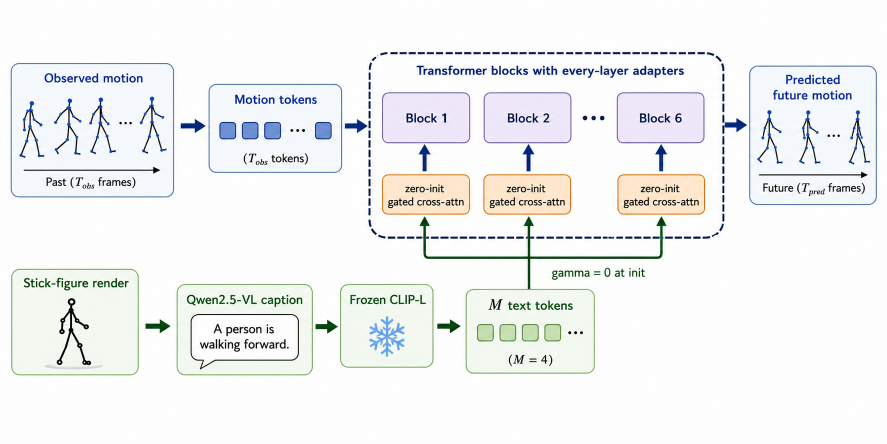}
  \Description{A two-stream architecture diagram. Observed motion is
  converted into motion tokens and passed through Transformer blocks.
  A stick-figure render is captioned, encoded by frozen CLIP-L, and
  projected into text tokens. The text tokens feed zero-gated
  cross-attention adapters after each Transformer block before future
  motion is predicted.}
  \caption{\textbf{Overview.} A VLM-generated caption is encoded by a
  frozen plain CLIP-L tower into a single 768-d vector. A learnable
  projection produces $M$ conditioning tokens. After every encoder
  block of the Transformer motion backbone, a zero-gated
  cross-attention adapter lets motion tokens attend to the
  conditioning tokens. Classifier-free dropout replaces a fraction of
  samples' text with a learnable null token during training. At
  initialisation the gate $\gamma$ is zero, so the model is
  numerically identical to the no-text baseline; any improvement is
  attributable to the adapter learning to use the text.}
  \label{fig:overview}
\end{figure*}

Pure motion sequences, however, are a narrow description of human
behavior. Many future motions are easier to interpret once the observed
motion is described at a semantic level: a person may be walking forward,
turning, sitting down, or preparing to reach. Modern vision-language
models (VLMs) can produce such descriptions from simple stick-figure
renders, turning each motion clip into a lightweight natural-language
prior. Text-to-motion generation has already shown the value of
language as a motion-level control signal~\cite{zhang2022motiondiffuse,
tevet2022mdm,zhang2023t2mgpt,guo2024momask,kim2023flame,
hong2025salad}. In prediction, the role of text is different: the
observed motion remains the primary evidence, while language should
quietly supply high-level context only when it helps.

This distinction motivates a design principle: text conditioning for
HMP should be useful but hard to overfit. The observed motion sequence
already carries most of the predictive signal, so language should serve
as a complementary cue rather than a replacement for motion dynamics.
For a well-tuned transformer predictor, it is natural to ask whether
semantic captions can be used through a small and transparent interface.

We propose \textsc{ZGL}, a text-guided zero-gated
cross-attention recipe for 3D HMP. Starting from a DCT-based
spatial-temporal Transformer backbone, we encode a VLM-generated
caption with a frozen plain CLIP-L text tower, project the caption
embedding into a small set of conditioning tokens, and insert a
cross-attention adapter after each motion encoder block.
The critical choice is zero-gating: every adapter output is multiplied
by a learnable gate initialised to zero~\cite{alayrac2022flamingo,
peebles2023dit}. At initialization, the full model is numerically
identical to the no-text baseline. During training, the adapter can
learn to use text only where the caption supplies predictive
information.

The contributions of this paper are:
\begin{itemize}
  \item We introduce \textsc{ZGL}, a compact text-conditioning
        module for 3D human motion prediction that adds zero-gated
        cross-attention adapters to a strong DCT-based
        spatial-temporal Transformer backbone.
  \item We show that VLM-generated captions can improve a competitive
        Human3.6M predictor with a simple frozen CLIP-L text tower and
        a small set of conditioning tokens.
  \item We validate the same recipe on CMU-Mocap, showing that the
        method remains competitive when transferred to a second
        benchmark.
\end{itemize}

\section{Method}
\label{sec:method}

\subsection{Backbone}

We take a DCT-based spatial-temporal Transformer baseline as our
motion backbone~\cite{aksan2021spatio,guo2023back}.
Given an observed window $x \in \mathbb{R}^{B \times T_{\text{in}} \times J \times 3}$
of $T_{\text{in}}\!=\!10$ subsampled frames and $J$ joints (22 for
Human3.6M and 25 for CMU), the input is first transformed
into the discrete cosine transform (DCT) domain along the time axis
following siMLPe~\cite{guo2023back}. A linear joint embedding lifts
the per-joint coordinates to $C\!=\!128$ channels and a learnable
positional embedding is added.

The backbone is an $N\!=\!6$-block stack. Each block runs a parallel
spatial attention and temporal attention, fused by a learned soft
mixture. We re-use the published hyperparameters: 8 heads, MLP ratio 4,
drop-path 0.2,
LayerScale init $10^{-5}$, learning rate $5\!\times\!10^{-4}$ with
LR decay 0.999, batch size 64, and 58 epochs of
training.  The output head is a two-layer MLP that maps the encoded
sequence back to joint coordinates, followed by inverse DCT and a
residual on the last input frame as in~\cite{guo2023back}.

\begin{table*}[!t]
  \centering
  \caption{\textbf{State-of-the-art comparison on Human3.6M.} MPJPE (mm); lower is better.}
  \label{tab:sota_h36m}
  \resizebox{\textwidth}{!}{%
  \begin{tabular}{l|cccc|cc|ccc}
    \hline
    \multirow{2}{*}{Method}
    & \multicolumn{4}{c|}{Short-term (ms)}
    & \multicolumn{2}{c|}{Long-term (ms)}
    & \multicolumn{3}{c}{Average} \\
    \cline{2-10}
    & 80 & 160 & 320 & 400 & 560 & 1000
    & Short & Long & Overall \\
    \hline
    Res-sup~\cite{martinez2017human}
    & 34.70 & 62.00 & 101.10 & 115.50 & 97.60 & 130.50
    & 78.33 & 114.05 & 90.23 \\
    DMGNN~\cite{li2020dynamic}
    & 17.00 & 33.60 & 65.90 & 79.70 & 103.00 & 137.20
    & 49.05 & 120.10 & 72.73 \\
    LTD/Traj-GCN~\cite{mao2019learning}
    & 12.70 & 26.10 & 52.30 & 63.50 & 81.60 & 114.30
    & 38.65 & 97.95 & 58.42 \\
    MSR-GCN~\cite{dang2021msr}
    & 12.10 & 25.60 & 51.60 & 62.90 & 81.10 & 114.20
    & 38.05 & 97.65 & 57.92 \\
    DSTD-GC~\cite{fu2023dstd}
    & 10.38 & 23.34 & 48.77 & 59.84 & 77.81 & 111.02
    & 35.58 & 94.42 & 55.19 \\
    DpNet~\cite{tang2023dpnet}
    & 10.30 & 22.90 & 47.90 & 58.10 & 77.00 & 109.90
    & 34.80 & 93.45 & 54.35 \\
    GuidanceReg~\cite{du2024guidancereg}
    & 10.60 & 22.80 & 47.30 & 58.30 & 76.60 & \textbf{109.60}
    & 34.75 & 93.10 & 54.20 \\
    PGBIG~\cite{ma2022progressively}
    & 10.30 & 22.70 & 47.40 & 58.50 & 76.90 & 110.30
    & 34.73 & 93.60 & 54.35 \\
    \textsc{ZGL} (Ours)
    & \textbf{9.27} & \textbf{20.60} & \textbf{45.14} & \textbf{56.32}
    & \textbf{75.15} & 109.96
    & \textbf{32.83} & \textbf{92.56} & \textbf{52.74} \\
    \hline
  \end{tabular}}
\end{table*}

\begin{table*}[!t]
  \centering
  \begin{minipage}[t]{0.49\textwidth}
  \centering
  \caption{\textbf{CMU-Mocap SOTA comparison.} MPJPE (mm).}
  \label{tab:sota_cmu}
  \scriptsize
  \setlength{\tabcolsep}{2.5pt}
  \resizebox{\linewidth}{!}{%
  \begin{tabular}{@{}lrrrrrr@{}}
    \toprule
    Method & 80 & 160 & 320 & 400 & 1000 & Avg.\\
    \midrule
    LTD~\cite{mao2019learning}      & 8.10 & 17.70 & 33.00 & 40.90 & 86.20 & 37.18 \\
    DMGNN~\cite{li2020dynamic}      & 13.60 & 24.10 & 47.00 & 58.80 & 112.60 & 51.22 \\
    MSR-GCN~\cite{dang2021msr}      & 8.10 & 15.20 & 30.60 & 38.60 & 83.00 & 35.10 \\
    STSGCN~\cite{sofianos2021space} & 10.80 & 18.40 & 31.20 & 41.20 & 81.70 & 36.66 \\
    PGBIG~\cite{ma2022progressively}& 7.60 & 14.30 & 29.00 & 36.60 & 81.70 & 33.84 \\
    SPGSN~\cite{li2022skeleton}     & 8.30 & 14.80 & 28.70 & 37.00 & \textbf{77.80} & 33.32 \\
    DpNet~\cite{tang2023dpnet}      & 8.40 & 14.50 & 30.50 & 39.70 & 91.30 & 36.88 \\
    \textsc{ZGL} (Ours)        & \textbf{6.57} & \textbf{11.96} & \textbf{25.78} & \textbf{33.38} & 83.17 & \textbf{32.17} \\
    \bottomrule
  \end{tabular}}
\end{minipage}

  \hfill
  \begin{minipage}[t]{0.47\textwidth}
  \centering
  \caption{\textbf{Injection placement ablation.} MPJPE (mm); lower is better.}
  \label{tab:abl_layer}
  \scriptsize
  \setlength{\tabcolsep}{2.5pt}
  \resizebox{\linewidth}{!}{%
  \begin{tabular}{@{}lrrrrc@{}}
    \toprule
    Layers & 80 & 160 & 320 & 400 & short-AVG\\
    \midrule
    1       & 9.58 & 21.08 & 45.87 & 57.21 & 33.44 \\
    1,2     & 9.52 & 20.98 & 46.05 & 57.39 & 33.48 \\
    1,2,3   & 9.68 & 21.16 & 46.01 & 57.36 & 33.55 \\
    1,2,3,4 & 9.48 & 21.00 & 45.88 & 57.36 & 33.43 \\
    1,2,3,4,5 & 9.47 & 21.03 & 45.94 & 57.21 & 33.41 \\
    all & \textbf{9.27} & \textbf{20.60} & \textbf{45.14} & \textbf{56.32} & \textbf{32.83} \\
    \bottomrule
  \end{tabular}}
\end{minipage}

\end{table*}

\subsection{Text encoder and projection}

Each input window is described by a one-sentence caption generated by
Qwen2.5-VL applied to a stick-figure rendering of the observed motion
only; future frames are not used for caption generation. The caption is
encoded by a frozen CLIP-ViT-L/14 text tower into a single
$d_\text{text}\!=\!768$-dimensional vector $t \in \mathbb{R}^{d_\text{text}}$.
We use the standard 77-token CLIP encoder; the impact of this choice
relative to BGE-M3 and Long-CLIP-L is measured in
Section~\ref{sec:exp:ablations}.
The text tower is never fine-tuned. This keeps the caption pathway
separate from motion learning and avoids introducing a second large
optimization problem into the motion predictor. It also makes the
conditioning signal cheap to cache: each training window only needs one
caption embedding, while all temporal reasoning remains inside the
motion backbone.

A small projection head maps $t$ into $M$ conditioning tokens
\begin{equation}
  \mathbf{t}_{\text{kv}} = \texttt{LN}\bigl(\texttt{GELU}(W\,t + b)\bigr)
  \in \mathbb{R}^{M \times C},
\end{equation}
with a learnable null token $t_\emptyset \in \mathbb{R}^{M \times C}$
initialised to zero.  Following classifier-free
guidance~\cite{ho2022classifier}, during training we replace
$\mathbf{t}_{\text{kv}}$ by $t_\emptyset$ for a fraction $p$ of
samples (we use $p\!=\!0.1$), so the model also learns an
unconditional mode and cannot collapse onto the text channel alone.

\subsection{Zero-gated cross-attention adapter}

Our method, \textsc{ZGL}, inserts a text-guided zero-gated
cross-attention adapter after every encoder block
$\ell\!\in\!\{0,\ldots,N{-}1\}$. The
intermediate representation $x_\ell \in \mathbb{R}^{B \times T \times J \times C}$
is augmented by an additive cross-attention adapter:
\begin{align}
  q   &= \texttt{LN}_q(x_\ell) \in \mathbb{R}^{B \times TJ \times C},\\
  k,v &= \texttt{LN}_{kv}(\mathbf{t}_{\text{kv}}) \in \mathbb{R}^{B \times M \times C},\\
  a   &= \texttt{MHA}(q, k, v),\\
  x_{\ell+1} &= x_\ell + \gamma \odot \texttt{Proj}(a),
  \label{eq:adapter}
\end{align}
where $\texttt{MHA}$ is multi-head attention with 8 heads,
$\texttt{Proj}$ is a $C \to C$ linear with dropout, and
$\gamma \in \mathbb{R}^{C}$ is a per-channel learnable gate
initialised to zero~\cite{touvron2021going, alayrac2022flamingo,
peebles2023dit, chen2023pixart}.  At initialisation
$\gamma = \mathbf{0}$ so the
adapter is an exact identity and the model is numerically equivalent
to the no-text Transformer baseline (Figure~\ref{fig:overview}). The
adapter has $\approx 200\text{k}$ parameters per layer, increasing
the model size from 4.5\,M to 5.7\,M parameters --
less than a 27\,\% addition on top of the baseline.
This identity start is important in HMP because the pose stream is
already highly predictive. A non-zero text branch can
disturb a well-trained motion prior before the caption features become
useful. In contrast, the zero gate lets the optimizer first recover the
pose-only solution and then open individual adapter channels only when
cross-attention consistently reduces prediction error.

The set of layers that get an adapter is a hyperparameter,
$\mathcal{I}\!\subseteq\!\{0,\ldots,N{-}1\}$; we treat
$\mathcal{I}\!=\!\{0,\ldots,N{-}1\}$ (every layer) as our default
and ablate over five subsets in Section~\ref{sec:exp:ablations}.

\subsection{Training objective}

The training loss is the joint position MPJPE on the output sequence
of length $T_{\text{out}}\!=\!10$ plus three auxiliary terms from
siMLPe. These auxiliary terms are the 3D velocity, bone-length, and
mutual-information losses, with weights as published
in~\cite{guo2023back}.  We also keep mirror augmentation,
time-reversal augmentation, and sample rate 2 identical to the no-text
baseline.

\begin{table*}[!t]
  \centering
  \begin{minipage}[t]{0.49\textwidth}
  \centering
  \caption{\textbf{Conditioning-source ablation.} MPJPE (mm); lower is better.}
  \label{tab:main}
  \scriptsize
  \setlength{\tabcolsep}{2.5pt}
  \resizebox{\linewidth}{!}{%
  \begin{tabular}{@{}lcrrrrc@{}}
    \toprule
    Source & dim & 80 & 160 & 320 & 400 & short-AVG\\
    \midrule
    Random & 768  & 9.79 & 21.81 & 47.75 & 59.30 & 34.66 \\
    Zero   & 768  & 9.78 & 21.36 & 46.45 & 57.88 & 33.87 \\
    \midrule
    No text & --   & 9.55 & 21.36 & 46.68 & 58.11 & 33.92 \\
    \midrule
    BGE-M3      & 1024 & 9.52 & 21.10 & 46.04 & 57.30 & 33.49 \\
    Long-CLIP-L & 768  & 9.58 & 21.16 & 46.04 & 57.27 & 33.51 \\
    CLIP-L      & 768  & \textbf{9.27} & \textbf{20.60} & \textbf{45.14} & \textbf{56.32} & \textbf{32.83} \\
    \bottomrule
  \end{tabular}}
\end{minipage}

  \hfill
  \begin{minipage}[t]{0.47\textwidth}
  \centering
  \caption{\textbf{Conditioning-token count ablation.} MPJPE (mm); lower is better.}
  \label{tab:abl_M}
  \scriptsize
  \setlength{\tabcolsep}{4pt}
  \resizebox{\linewidth}{!}{%
  \begin{tabular}{@{}rrrrrc@{}}
    \toprule
    $M$ & 80 & 160 & 320 & 400 & short-AVG\\
    \midrule
    1 & 9.45 & 21.04 & 45.85 & 57.09 & 33.36 \\
    2 & 9.48 & 21.00 & 46.15 & 57.53 & 33.54 \\
    4 & \textbf{9.27} & \textbf{20.60} & \textbf{45.14} & \textbf{56.32} & \textbf{32.83} \\
    8 & 9.48 & 20.94 & 45.92 & 57.26 & 33.40 \\
    \bottomrule
  \end{tabular}}
\end{minipage}

\end{table*}

\section{Experiments}
\label{sec:exp}

\subsection{Datasets and evaluation protocol}
\label{sec:exp:setup}

\paragraph{Datasets.}
We evaluate on Human3.6M~\cite{ionescu2014h36m} and CMU Mocap. We
follow the standard Human3.6M split with S1, S6, S7, S8, and S9 for
training, S11 for validation, and S5 for testing on 22 joints. CMU uses
the common 25-joint evaluation setting adopted by recent HMP studies.

\paragraph{Metric.}
We observe $T_{\text{in}}\!=\!10$ frames and report MPJPE at the
dataset-standard horizons: 80, 160, 320, 400, 560, and 1000~ms on
Human3.6M and CMU Mocap. We also report short-AVG, long-AVG, and
Total-AVG where applicable. We select the epoch with the best
validation short-AVG.

\paragraph{Text features.}
Qwen2.5-VL captions from stick-figure renders are encoded by frozen
BGE-M3~\cite{chen2024bge}, Long-CLIP-L~\cite{zhang2024longclip}, or
plain CLIP-ViT-L/14~\cite{radford2021clip}; plain CLIP-L is default.

\subsection{State-of-the-art comparisons}
\label{sec:exp:main}

Table~\ref{tab:sota_h36m} compares \textsc{ZGL} with
representative Human3.6M state-of-the-art methods using plain CLIP-L
captions, $M\!=\!4$, and every-layer injection. We select our model by
validation short-AVG. \textsc{ZGL}
achieves the lowest short-AVG, the lowest 560~ms error, and the best
overall average in this comparison.
The gains are concentrated in the short and middle horizons, where the
observed motion still constrains the future and semantic intent can
disambiguate similar local dynamics. At 1000~ms the method remains
competitive, which is consistent with captions acting as high-level
context rather than replacing long-range motion dynamics.

Table~\ref{tab:sota_cmu} compares \textsc{ZGL} with
representative CMU-Mocap methods at 80, 160, 320, 400, and 1000~ms.
For CMU, we reuse the Human3.6M architecture and hyperparameter
setting: plain CLIP-L captions, $M\!=\!4$, every-layer adapters, and
dropout probability $p\!=\!0.1$. Under this setting,
\textsc{ZGL} obtains the best average MPJPE and the lowest error
at the 80--400~ms horizons, while SPGSN remains stronger at 1000~ms.

\subsection{Ablation studies}
\label{sec:exp:ablations}

We vary one design axis at a time around the Table~\ref{tab:sota_h36m}
recipe: plain CLIP-L, $M\!=\!4$, every-layer injection, and dropout
$p\!=\!0.1$.

\paragraph{Conditioning source.}
Table~\ref{tab:main} compares alternative real-text encoders with
deterministic random and all-zero controls. The SOTA plain CLIP-L
setting gives the best short-AVG, while random vectors degrade
performance, indicating that the gain is not explained by extra adapter
parameters alone.
The all-zero row is close to the no-text baseline, as expected from the
zero-gated initialization: without semantic variation in the text
stream, the adapter has little useful signal to exploit. The random row
is worse, showing that arbitrary conditioning vectors can make the
adapter attend to non-semantic noise.

\paragraph{Injection placement.}
Table~\ref{tab:abl_layer} tests whether text should be fused late by
restricting adapters to different encoder-block subsets.
The prefix scan keeps all other settings fixed and gradually exposes
more motion blocks to the caption tokens. This isolates whether the
caption should act only after motion features are deep, or whether
earlier spatial-temporal representations can also benefit from the
semantic prior.
The all-layer setting performs best, but the prefix variants remain
close, suggesting that text does not need a deep specialized branch but
should be available throughout the motion stack.

\paragraph{Token capacity.}
Table~\ref{tab:abl_M} varies the number of conditioning tokens
$M\!\in\!\{1,2,4,8\}$.
The adapter uses these tokens only as keys and values, so increasing
$M$ expands the text memory seen by motion queries without changing the
motion-token length. The best setting is $M\!=\!4$, suggesting that one
global caption vector is too compressed, while a larger token budget
does not add useful information for the short captions used here.

\section{Conclusion}
\label{sec:conclusion}

We presented \textsc{ZGL}, a simple recipe for text-conditioned 3D
human motion prediction: a zero-gated cross-attention adapter after
every Transformer encoder block, conditioned by tokens projected from a
frozen plain CLIP-L caption embedding. The method improves the no-text
baseline on Human3.6M with a lightweight adapter pathway.
The main finding is that language does not need to dominate the motion
model to be useful. When captions are generated from the observed
motion only and injected through zero-initialized gates, the backbone
can keep its original dynamics at the start of training and gradually
learn where semantic context helps. The ablations support this view:
real caption features outperform random or zero controls, all-layer
injection is more effective than late-only fusion, and a small set of
conditioning tokens is sufficient for concise motion descriptions.

These results suggest a practical direction for language-assisted HMP.
Instead of building a separate text-heavy prediction system, a strong
DCT-based spatial-temporal Transformer can be extended with a compact
semantic interface that is easy to train and inexpensive at inference
once captions are precomputed. Future work can study richer caption
generation, online visual observations, and more explicit handling of
ambiguous future motion while keeping the motion backbone as the
central predictor.

\begin{acks}
This work was supported in part by Hunan Key Research and Development
Program under Grant 2024WK2005.
\end{acks}

\clearpage
\bibliographystyle{ACM-Reference-Format}
\bibliography{references}

\end{document}